%% file: main.tex
\documentclass[letterpaper]{article} 
\usepackage{aaai2026}  
\usepackage{times}  
\usepackage{helvet}  
\usepackage{courier}  
\usepackage[hyphens]{url}  
\usepackage{graphicx} 
\usepackage{natbib}  
\usepackage{caption} 
\usepackage{algorithm}
\usepackage{algorithmic}

\usepackage{booktabs}       
\usepackage{amsfonts}       
\usepackage{nicefrac}       
\usepackage{microtype}      
\usepackage{xcolor}         

\usepackage{xspace} 
\usepackage{amsmath}
\usepackage{amssymb}
\usepackage{multirow}
\usepackage{multicol}
\usepackage{graphicx}

\usepackage{pifont}

\usepackage{float}  
\usepackage{tikz}
\usepackage{pgfplots}
\pgfplotsset{compat=1.18}
\usepackage{subcaption}
\usepackage[capitalise]{cleveref}
\usepackage[table,dvipsnames]{xcolor}

\usepackage[english]{babel}

\usepackage{newfloat}
\usepackage{listings}
\DeclareCaptionStyle{ruled}{labelfont=normalfont,labelsep=colon,strut=off} 
\floatstyle{ruled}
\newfloat{listing}{tb}{lst}{}
\floatname{listing}{Listing}
\title{Boosting Resolution Generalization of Diffusion Transformers with\\Randomized Positional Encodings}
\author{
    Liang Hou\textsuperscript{\rm 1}\equalcontrib, Cong Liu\textsuperscript{\rm 1,2}\equalcontrib\thanks{This work was conducted during the author's internship at Kling Team, Kuaishou Technology.}, Mingwu Zheng\textsuperscript{\rm 1}, Xin Tao\textsuperscript{\rm 1}\thanks{Corresponding author.},
    Pengfei Wan\textsuperscript{\rm 1},
    Di Zhang\textsuperscript{\rm 1},
    Kun Gai\textsuperscript{\rm 1}
}
\affiliations{
    \textsuperscript{\rm 1}Kling Team, Kuaishou Technology\\
    \textsuperscript{\rm 2}Southeast University\\

    \{lianghou96, jiangsutx\}@gmail.com \{liucong08, zhengmingwu, wanpengfei, zhangdi08\}@kuaishou.com
}

\usepackage{bibentry}

\begin{document}

\maketitle

\begin{abstract}
Resolution generalization in image generation tasks enables the production of higher-resolution images with lower training resolution overhead. However, a key obstacle for diffusion transformers in addressing this problem is the mismatch between positional encodings seen at inference and those used during training. Existing strategies such as positional encodings interpolation, extrapolation, or hybrids, do not fully resolve this mismatch. In this paper, we propose a novel two-dimensional randomized positional encodings, namely RPE-2D, that prioritizes the order of image patches rather than their absolute distances, enabling seamless high- and low-resolution generation without training on multiple resolutions. Concretely, RPE-2D independently samples positions along the horizontal and vertical axes over an expanded range during training, ensuring that the encodings used at inference lie within the training distribution and thereby improving resolution generalization. We further introduce a simple random resize-and-crop augmentation to strengthen order modeling and add micro-conditioning to indicate the applied cropping pattern. On the ImageNet dataset, RPE-2D achieves state-of-the-art resolution generalization performance, outperforming competitive methods when trained at $256^2$ and evaluated at $384^2$ and $512^2$, and when trained at  $512^2$ and evaluated at $768^2$ and $1024^2$. RPE-2D also exhibits outstanding capabilities in low-resolution image generation, multi-stage training acceleration, and multi-resolution inheritance.
\end{abstract}

\input{sections/1_introduction}

\input{sections/2_related_work}

\input{sections/3_preliminaries}

\input{sections/4_method}

\input{sections/5_experiments}

\input{sections/6_conclusion}

\bibliography{main}

\end{document}

%% file: sections/1_introduction.tex
\section{Introduction}

Diffusion models~\cite{ho2020denoising,nichol2021improved,song2020denoising,song2020score} have effectively replaced traditional generative models such as variational autoencoders~\cite{kingma2013auto} and generative adversarial networks~\cite{goodfellow2014generative} as the predominant paradigm in the field of image generation due to their strong generative performance~\cite{dhariwal2021diffusion,rombach2022high}.
Diffusion Transformers (DiTs)~\cite{peebles2023scalable} further demonstrate that Transformers~\cite{vaswani2017attention} can be effectively scaled within diffusion frameworks, making Transformer-based diffusion architectures one of the central focuses of modern diffusion model research~\cite{lu2024fit,ma2024sit,chen2023pixart,chen2024gentron}.
However, existing image generation models are typically trained at a specific resolution to produce high-quality images only at that resolution. Scaling these models directly to higher resolutions usually incurs a multiplicative increase in training cost, which becomes prohibitive when computation and data resources are limited. This situation calls for models with genuine cross-resolution generalization ability, such that, even when trained solely on low-resolution images, they can still generate high-quality images at higher resolutions, thereby avoiding the substantial costs associated with conventional high-resolution training.

A number of approaches have been proposed to address resolution generalization of image generation. The first line of work~\cite{he2023scalecrafter,du2024demofusion,lu2024fit,teterwak2019boundless,yang2019very} focuses on enhancing network architectures, but often leads to complex designs that are tightly coupled to specific frameworks or training pipelines. A second line of work~\cite{jin2023training} improves extrapolation by modifying the attention mechanism to account for changes in attention entropy. However, it largely overlooks a key bottleneck: the one-to-one correspondence introduced by positional encodings (PEs), which enables Transformers to perceive positional information but simultaneously constrains the resolution generalization capacity of DiTs. A third line of methods~\cite{zhuo2024lumina,lu2024fit,peng2023yarn,NTK-aware} explicitly targets the limitations that PEs impose on generalization, proposing interpolation-, extrapolation-, or hybrid-based schemes. Yet these methods remain bounded by the intrinsic extrapolation limits of the underlying PEs and do not fully close the PE gap between training and inference.

In this work, we revisit resolution generalization in image generation from the perspective of PEs. We argue that the fundamental reason existing methods perform poorly at resolution extrapolation is that many PEs required at test time have never “truly” appeared during training, leading to a systematic distributional mismatch of PEs between training and inference. To fundamentally alleviate this issue, we posit that all PEs used at test time should, in a statistical sense, be “covered” by the sampling process during training. Guided by this principle, and inspired by the success of one-dimensional randomized positional encodings (RPE-1D) in handling length extrapolation in large language models~\cite{ruoss2023randomized}, we propose RPE-2D, a two-dimensional, training-based randomized positional encoding framework tailored for resolution generalization in image generation.

In contrast to conventional approaches that attempt to extend positions along fixed coordinate axes, RPE-2D performs random sampling over a larger two-dimensional grid while only enforcing consistency of order along the horizontal and vertical axes. As a result, all PEs required during high-resolution inference can be regarded as statistically covered by the random sampling process at training time. This reframes an out-of-domain extrapolation task as an in-domain interpolation problem and models them in a unified manner via random selection, thereby avoiding any additional training overhead. Conceptually, each image can be regarded as a cropped, resized, or geometrically transformed view of a larger latent canvas. This is fundamentally different from the one-dimensional textual sequences processed by language models, where it is natural to assume a uniform step size between adjacent tokens along the sequence and to use equally spaced positional encodings to represent their order. In contrast, in two-dimensional visual settings, different views correspond to different regions and scales of the same underlying canvas. From this perspective, using exactly the same, equally spaced positional encodings to model all such views introduces unnecessary constraints on positional modeling. The design of RPE-2D is precisely motivated by this observation: by assigning randomized two-dimensional positional encodings, it aims to weaken the model's reliance on specific positional intervals and instead encourage it to exploit positional order, which is an essential factor that has often been overlooked in prior work.

Concretely, RPE-2D performs without-replacement random sampling along the horizontal and vertical axes of a predefined maximal grid, followed by sorting the sampled indices in ascending order to construct a two-dimensional set of random positions. At test time, we instead adopt a deterministic, equidistant sampling strategy to achieve better generalization in expectation. To further enhance the model’s ability to capture positional order, we introduce a data augmentation strategy that combines random resizing and cropping, and employ micro-conditioning to explicitly inject the corresponding cropping and resizing information. This allows the model to preserve the topological structure of images while relying more on positional order than on precise distances. In addition, we incorporate attention scaling and timestep shifting strategies during inference to alleviate performance degradation caused by changes in attention entropy and signal-to-noise ratio when sampling at high resolutions.

We empirically validate RPE-2D on ImageNet at both $256^2$ and $512^2$ training resolutions. When trained at $256^2$ and evaluated at $384^2$ and $512^2$, as well as trained at $512^2$ and evaluated at $768^2$ and $1024^2$, RPE-2D consistently outperforms strong positional-encoding extrapolation baselines, demonstrating state-of-the-art resolution generalization performance under all evaluation settings. Moreover, integrating RPE-2D with different PE families maintains or improves in-distribution image quality at the training resolution, indicating that RPE-2D is broadly compatible with existing DiT architectures. Beyond upward resolution extrapolation, RPE-2D also supports downward resolution generation, accelerates multi-stage training when fine-tuning to higher resolutions, and enables flexible multi-resolution inheritance, highlighting its practical value for scalable diffusion transformers.

%% file: sections/2_related_work.tex
\section{Related Work}

\subsection{Length Generalization in Languages Models}
A significant stride in extrapolation has been achieved with ALIBI~\cite{press2021train}, a method that employs local attention to reinforce the model's ability to capture local dependencies within the data. This is crucial as it allows the model to maintain a more refined understanding of the data's structure, thereby improving the quality of extrapolation.Another notable approach is the NTK~\cite{NTK-aware}, which adjusts the frequency components of the position encodings. This method is designed to preserve the high-frequency information during the extrapolation process, ensuring a more accurate representation of the data's characteristics.
YaRN~\cite{peng2023yarn} is an innovative approach that extends the context window of large language models efficiently. It does so by modifying the attention mechanism to handle longer sequences without the need for fine-tuning, thus maintaining a consistent level of performance across various lengths of input data.
The concept of random position encoding~\cite{ruoss2023randomized} has also gained traction, offering a more natural and elegant solution to the challenge of handling longer sequences during prediction. This method has been shown to be effective not only in language models but also in non-language models, where the generation of images or other data types requires a broader context understanding.
Attention Masking is another strategy that has proven effective in language models, which are inherently local in nature. By "forcing" the model to focus on a limited number of tokens, it can effectively manage the increased complexity during prediction. However, its applicability to non-language models is still under exploration .

\subsection{Resolution Generalization in Diffusion Models}
In the realm of computer vision, extrapolation techniques have been pivotal in advancing the capabilities of models to generate images and predict video sequences beyond the limits of their training data. The development of FiT~\cite{lu2024fit} and LuminaNext~\cite{zhuo2024lumina} has showcased the potential of local attention mechanisms in enhancing the performance of image generation models. Local attention focuses on specific regions within an image, allowing for more detailed and accurate generation of high-resolution images. 
In addition to these, there are modifications to the network structure, such as attention scale~\cite{jin2023training}, neighborhood attention~\cite{hassani2023neighborhood}, and KV-compression~\cite{chen2024pixart}.
In summary, while current methods have made limited improvements in extrapolation capabilities, they still fail to address the fundamental issue of the position encoding gap between training and prediction.

%% file: sections/3_preliminaries.tex
\begin{figure*}[!t]
    \captionsetup{type=figure}
    \includegraphics[width=1.0\linewidth]{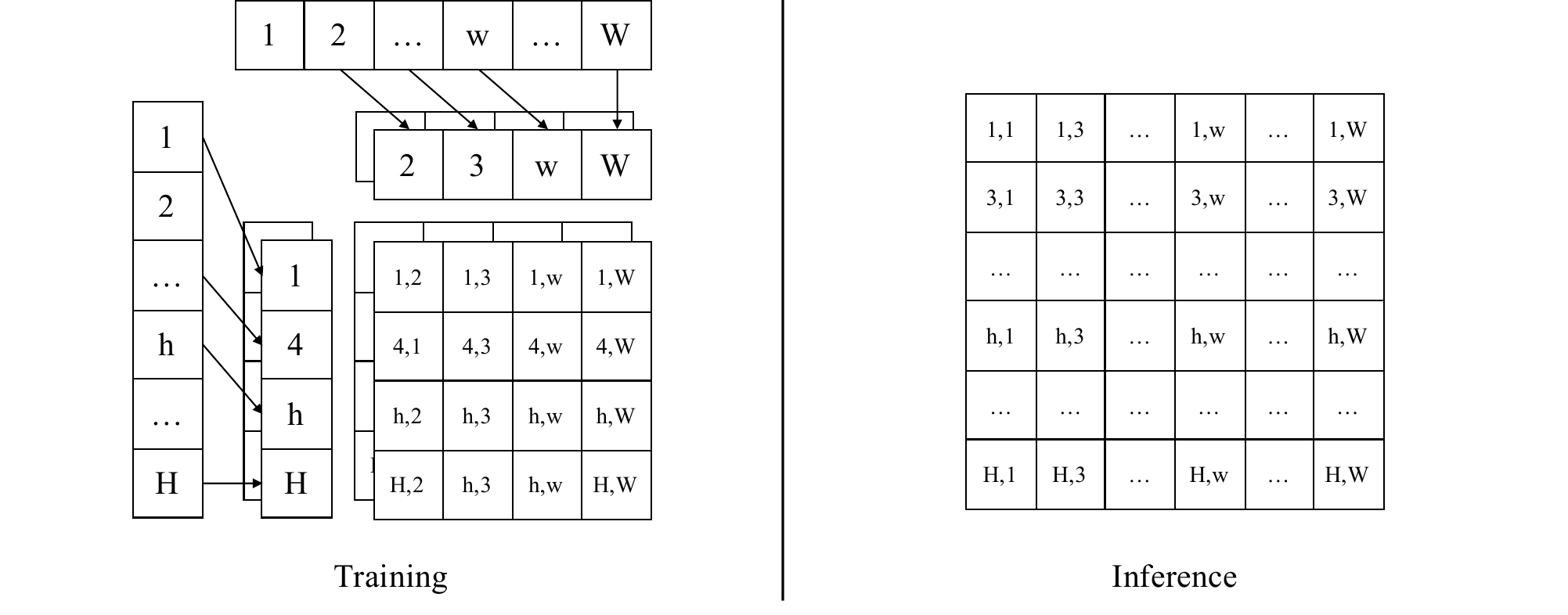}
    \caption{Illustration of RPE-2D for training and inference. During training (left), row and column indices are randomly sampled without replacement from the maximal grid $H \times W$ and sorted to form a set of 2D positions matching the training resolution. During inference (right), a deterministic, approximately equidistant grid matching the inference resolution is used.}
    \label{fig_rpe}
\end{figure*}

\section{Preliminary}

\subsection{Positional Encodings}

\paragraph{Sinusoidal PE}
Positional encodings (PEs)~\cite{vaswani2017attention} play a significant role in Transformer-based sequence modeling, as they inject positional information into token representations to compensate for the order-agnostic nature of self-attention.
A widely used choice is the sinusoidal PE, which adds to each token embedding $\mathbf{x}_m \in \mathbb{R}^d$ at position $m \in \{1,2,\dots,L\}$ a positional vector $\mathrm{PE}(m) := \mathbf{p}_m \in \mathbb{R}^d$, where $d \in \mathbb{N}^+$ is the embedding dimension.
Its components are defined as
\begin{align}
    \mathrm{PE}(m,2i)   &:= p_{m,2i}   = \sin(m \theta_i), \\
    \mathrm{PE}(m,2i+1) &:= p_{m,2i+1} = \cos(m \theta_i),
\end{align}
where $i \in \{0,1,\dots,d/2-1\}$ and
$\theta_i = b^{-2i/d}$ is the frequency associated with the $i$-th pair of dimensions, with base $b \in \mathbb{R}^+$.

\paragraph{RoPE}
Rotary positional encoding (RoPE)~\cite{su2024roformer} is a form of relative PE that has shown strong length generalization and has become a preferred choice in both modern LLMs and DiTs.
Instead of adding a positional vector, RoPE applies a position-dependent rotation to the query and key vectors in self-attention.
Let $\mathbf{q}_m \in \mathbb{R}^d$ and $\mathbf{k}_n \in \mathbb{R}^d$ denote the query and key at positions $m$ and $n$, respectively, and let $f$ denote the attention function.
RoPE modifies $f$ as
\begin{align}
    f(\mathbf{q}_m, \mathbf{k}_n, m, n)
    &= (\mathbf{R}_m \mathbf{q}_m)^\top (\mathbf{R}_n \mathbf{k}_n) \nonumber \\
    &= \mathbf{q}_m^\top \mathbf{R}_m^\top \mathbf{R}_n \mathbf{k}_n
     = \mathbf{q}_m^\top \mathbf{R}_{n-m} \mathbf{k}_n,
\end{align}
where $\mathbf{R}_m$ and $\mathbf{R}_n$ are rotation matrices that depend on the absolute positions, and
$\mathbf{R}_{n-m} := \mathbf{R}_m^\top \mathbf{R}_n$ depends only on the relative offset $(n-m)$.
For a single 2D subspace (a pair of channels), the relative rotation matrix takes the form
\begin{align}
\mathbf{R}_{n-m} =
\begin{pmatrix}
\cos((n-m)\theta_i) & -\sin((n-m)\theta_i) \\
\sin((n-m)\theta_i) &  \cos((n-m)\theta_i)
\end{pmatrix},
\end{align}
The full $d$-dimensional rotation matrix is block-diagonal, composed of such $2 \times 2$ rotation blocks.

\subsection{2D Positional Encodings}

For image-like data with a two-dimensional structure, PEs are typically extended to 2D by composing two independent 1D PEs along the horizontal and vertical axes.
Taking 2D RoPE as an example, consider the query $\mathbf{q}_{x_1,y_1}$ at spatial position $(x_1, y_1)$ and the key $\mathbf{k}_{x_2,y_2}$ at position $(x_2, y_2)$.
The corresponding 2D rotary matrix can be written as
\begin{align}
\mathbf{R}_{x_2-x_1,\,y_2-y_1}
=
\begin{pmatrix}
\mathbf{R}_{x_2-x_1} & \mathbf{0} \\
\mathbf{0}            & \mathbf{R}_{y_2-y_1}
\end{pmatrix},
\label{eq:RoPE2D}
\end{align}
where $\mathbf{R}_{x_2-x_1}$ and $\mathbf{R}_{y_2-y_1}$ are 1D RoPE rotation matrices along the horizontal and vertical directions, respectively, and the full 2D rotation is realized as a block-diagonal composition of the two.
This construction naturally adapts RoPE to 2D grids while preserving its relative-position property along each axis.

%% file: sections/4_method.tex
\section{Method}
\label{sec3_method}

\subsection{2D Randomized Positional Encodings}

We consider resolution generalization in image generation, where a model is trained only at a low resolution due to computational constraints but is expected to generate images at higher resolutions at test time.
Let $h_\mathrm{train}, w_\mathrm{train} \in \mathbb{N}^+$ denote the spatial size of the training images (or VAE latents), and $h_\mathrm{test}, w_\mathrm{test} \in \mathbb{N}^+$ that of the test images, with $h_\mathrm{test} > h_\mathrm{train}$ and $w_\mathrm{test} > w_\mathrm{train}$.
Under such resolution extrapolation, many positional encodings required at test time inevitably lie outside the range seen during training.

NTK~\cite{NTK-aware} and YaRN~\cite{peng2023yarn} extend the usable context range by combining interpolation and extrapolation, but they do not resolve a fundamental issue: the positional encoding associated with each token differs between training and inference.
Inspired by one-dimensional randomized positional encodings (RPE-1D)~\cite{ruoss2023randomized} in LLMs, we reinterpret resolution extrapolation in image generation as an interpolation problem and propose \emph{2D Randomized Positional Encodings} (RPE-2D).
The core idea is to ensure that all positional encodings used at test time are statistically covered by the training-time sampling process.
By randomly assigning positions to image patches in a structured manner, every test-time position lies within the training distribution, thereby improving robustness to positional shifts.

As illustrated in~\cref{fig_rpe}, RPE-2D extends RPE-1D, originally designed for text, to a two-dimensional setting suitable for images.
A naive extension would be to flatten the $h_\mathrm{train} \times w_\mathrm{train}$ patches into a 1D sequence and sample positions from a longer 1D range of length $HW$, where $H > h_\mathrm{test} > h_\mathrm{train}$ and $W > w_\mathrm{test} > w_\mathrm{train}$ are hyperparameters.
However, such flattening ignores the inherent 2D structure of images and entangles horizontal and vertical neighbors in an unnatural way, leading to distorted distances along the two axes.
For 2D image data, the horizontal and vertical axes are naturally decoupled.
RPE-2D therefore performs independent randomized position sampling along each axis.
Formally, at each training step we sample, without replacement, index sets
$\mathcal{X} \subset \{1, 2, \dots, H\}$ and
$\mathcal{Y} \subset \{1, 2, \dots, W\}$ such that
$|\mathcal{X}| = h_\mathrm{train}$ and $|\mathcal{Y}| = w_\mathrm{train}$.
We then sort them in ascending order,
$\mathcal{X} = \{x_1, \dots, x_{h_\mathrm{train}}\}$ with
$x_1 < x_2 < \dots < x_{h_\mathrm{train}}$
and
$\mathcal{Y} = \{y_1, \dots, y_{w_\mathrm{train}}\}$ with
$y_1 < y_2 < \dots < y_{w_\mathrm{train}}$.
The 2D random position set is constructed via the Cartesian product
\begin{equation}
\mathcal{X} \times \mathcal{Y}
= \{(x, y) \mid x \in \mathcal{X},\, y \in \mathcal{Y}\}.
\label{Cartesian product}
\end{equation}
For the patch at training index $(i,j)$, where
$1 \le i \le h_\mathrm{train}$ and $1 \le j \le w_\mathrm{train}$,
its randomized positional encoding is defined as
\[
\operatorname{RPE}(i, j) := \operatorname{PE}(x_i, y_j) \in \mathbb{R}^d,
\]
with $(x_i, y_j) \in \mathcal{X} \times \mathcal{Y}$ and $\operatorname{PE}(\cdot,\cdot)$ denoting any 2D positional encoding function (e.g., SinPE or RoPE).
This construction preserves the monotonic order along each axis and induces a consistent 2D grid structure: along any fixed row, vertical coordinates are aligned, and along any fixed column, horizontal coordinates are aligned, while the actual intervals between sampled positions vary across training steps and thus prevent the model from memorizing specific lengths.

At test time, RPE-2D uses deterministic and approximately equidistant positions.
Given a maximal grid of size $H \times W$, we choose
$x_1 = 1$, $x_{h_\mathrm{test}} = H$ and
$y_1 = 1$, $y_{w_\mathrm{test}} = W$,
with spacings
$x_{i+1} - x_i = \big\lfloor H / h_\mathrm{test} \big\rfloor$ and
$y_{j+1} - y_j = \big\lfloor W / w_\mathrm{test} \big\rfloor$.
In this way, all test-time positions lie within the support of the randomized training positions while covering the full spatial extent of the maximal grid, effectively turning resolution extrapolation into interpolation over a shared 2D positional range.
Our RPE-2D training paradigm is orthogonal to the specific choice of positional encoding and can be applied on top of both sinusoidal PEs and RoPE; we empirically validate this compatibility in our experiments (see~\cref{table_comparsion_different_PEs}).

\subsection{Data Augmentation and Micro-Conditioning}
To further enhance the model’s ability to perceive the order of image patches, we jointly apply resize and crop operations to transform the ``collected'' high-resolution images into low-resolution inputs suitable for training. The resize operation encourages the model to capture global structure, while the crop operation prompts it to attend to local details. Importantly, the low-resolution images produced by these two operations are kept at the same spatial resolution.
To address the issue of image incompleteness introduced by cropping, we design a micro-conditioning mechanism. We first upsample each low-resolution image in the training set to a high resolution (if necessary) and record its base resolution as
\(\mathbf{c}_{\text{original}} = (h_{\text{original}}, w_{\text{original}})\).
During each training iteration, we then randomly select start and end coordinates from a set of cropping options (including the no-crop case corresponding to global resizing) and crop the base image accordingly, yielding crop coordinates
\(\mathbf{c}_{\text{crop}} = (c_{\text{top}}, c_{\text{left}}, c_{\text{down}}, c_{\text{right}})\).
The cropped region is subsequently resized to a target resolution
\(\mathbf{c}_{\text{resize}} = (h_{\text{target}}, w_{\text{target}})\),
where \(h_{\text{target}} \times w_{\text{target}}\) matches the desired training resolution. These three types of conditioning information are injected into the model via adaLN~\cite{xu2019understanding}.
Concretely, each component is independently embedded using Fourier feature encoding~\cite{tancik2020fourier}, and the resulting embeddings are concatenated into a single vector. We then add this vector to the DiT~\cite{peebles2023scalable} timestep embedding, thereby providing the model with explicit information about the original resolution, cropping pattern, and final resize configuration.

\subsection{Training-Free Sampling Strategy}
\paragraph{Attention Scale} In addition to the changes in PEs, resolution extrapolation inevitably leads to an increase in the number of image patches, creating another inconsistency between testing and training. Since attention is scale-dependent, this dependency arises from the fact that the entropy of attention changes as the number of patches increases~\cite{jin2023training}. We also attempt to use the proposed scaling factor to mitigate the variations in attention entropy,
\begin{align}
\operatorname{Attention}(\mathbf{Q}, \mathbf{K}, \mathbf{V})=\operatorname{softmax}\left(\frac{\log _{n} m}{\sqrt{d}} \mathbf{Q} \mathbf{K}^{\top}\right) \mathbf{V},
\end{align}
 where $m = h_\text{test}\times w_\text{test}$ and $n = h_\text{train}\times w_\text{train}$ represent the number of patches during testing and training, respectively.

\paragraph{Timestep Shift}
When generating large images with diffusion models, the increase in resolution leads to an increase in the signal-to-noise ratio (SNR) of the noise schedule used in training~\cite{hoogeboom2023simple}. Therefore, it is necessary to adjust the inference timestep spacing during sampling to maintain the SNR as much as possible. Specifically, We follow SD3~\cite{esser2024scaling} to map the time tep $t_n\in\{1,2,\dots,T\}$ for $n$ patches in training to the timestep $t_m\in\{1,2,\dots,T\}$ for $m$ patches in inference to approximate the same level of SNR,
\begin{align}
t_m = \left\lfloor\frac{\sqrt{\frac{m}{n}} \times \frac{t_n}{T}}{1+\left(\sqrt{\frac{m}{n}}-1\right) \times \frac{t_n}{T}}\right\rfloor \times T.
\end{align}

\begin{table*}[ht]
    \centering
    \begin{tabular}{lccccc|ccccc}
    \toprule
    \multirow{3}{*}{\centering Method} & \multicolumn{10}{c}{ImageNet $256 \times 256$} \\
    \cmidrule(lr){2-11} 
     & \multicolumn{5}{c|}{$384 \times 384$} & \multicolumn{5}{c}{$512 \times 512$} \\
     & FID$\downarrow$ & sFID$\downarrow$ & IS$\uparrow$ & Precision$\uparrow$ & Recall$\uparrow$ & FID$\downarrow$ & sFID$\downarrow$ & IS$\uparrow$ & Precision$\uparrow$ & Recall$\uparrow$\\
    \midrule
    PI & 18.87 & 41.59 & 260.97 & 0.8312 & 0.0602 & 30.64 & 57.76 & 159.72 & 0.676 & 0.055 \\
    Ext & 15.95 & 37.79 & 374.74 & 0.8970 & 0.0636 & 28.35 & 54.77 & 232.41 & 0.607 & 0.181\\
    NTK & 16.56 & 35.92 & 375.28 & 0.9203 & 0.0686 & 27.88 & 49.8 & 227.45 & 0.619 & 0.177\\
    YaRN & 16.97 & 26.08 & 264.34 & 0.7864 & 0.1050 & 19.13 & 34.31 & 253.16 & 0.749 & 0.151 \\
    RPE-2D & \textbf{15.63} & \textbf{14.40} & \textbf{385.67} & \textbf{0.9631} & \textbf{0.1174} & \textbf{17.95} & \textbf{18.23} & \textbf{348.99} & \textbf{0.849} & \textbf{0.181} \\
    \bottomrule
    \end{tabular}
    \caption{Comparison of RPE-2D with different methods on resolution extrapolation trained on ImageNet $256 \times 256$.}
    \label{tab:Extrapolation_384_512}
\end{table*}

\begin{table*}[ht]
    \centering
    \begin{tabular}{lccccc|ccccc}
    \toprule
    \multirow{3}{*}{\centering Method} & \multicolumn{10}{c}{ImageNet $512 \times 512$} \\
    \cmidrule(lr){2-11}
     & \multicolumn{5}{c|}{$768 \times 768$} & \multicolumn{5}{c}{$1024 \times 1024$} \\
     &FID$\downarrow$ & sFID$\downarrow$ & IS$\uparrow$ & Precision$\uparrow$ & Recall$\uparrow$ &FID$\downarrow$ & sFID$\downarrow$ & IS$\uparrow$ & Precision$\uparrow$ & Recall$\uparrow$ \\
    \midrule
    PI & 27.24 & 68.63 & 150.53 & \textbf{0.8441} & 0.373 & 38.64 & 92.49 & 139.38 & \textbf{0.7066} & 0.361\\
    Ext & 20.57 & 53.65 & 223.39 & 0.8184 & 0.462 & 45.77 & 116.54 & 180.16 & 0.5698 & 0.499 \\
    NTK & 21.58 & 46.11 & 225.00 & 0.7883 & 0.452 & 32.90 & 73.69 & \textbf{216.12} & 0.6304 & 0.579\\
    YaRN & 55.21 & 75.84 & 62.93 & 0.5547 & 0.513 & 50.65 & 79.04 & 103.38  & 0.6010 & 0.461\\
    RPE-2D & \textbf{20.45} & \textbf{40.46} & \textbf{271.05} & 0.8271 & \textbf{0.512} & \textbf{25.40} & \textbf{47.18} & 192.05 & 0.8109 & \textbf{0.535}\\
    \bottomrule
    \end{tabular}
    \caption{Comparison of RPE-2D with different methods on resolution extrapolation trained on ImageNet $512 \times 512$.}
    \label{tab:Extrapolation_768_1024}
\end{table*}

%% file: sections/5_experiments.tex
\section{Experiments}

\begin{figure*}[!t]
    \captionsetup{type=figure}
    \includegraphics[width=1.0\linewidth]{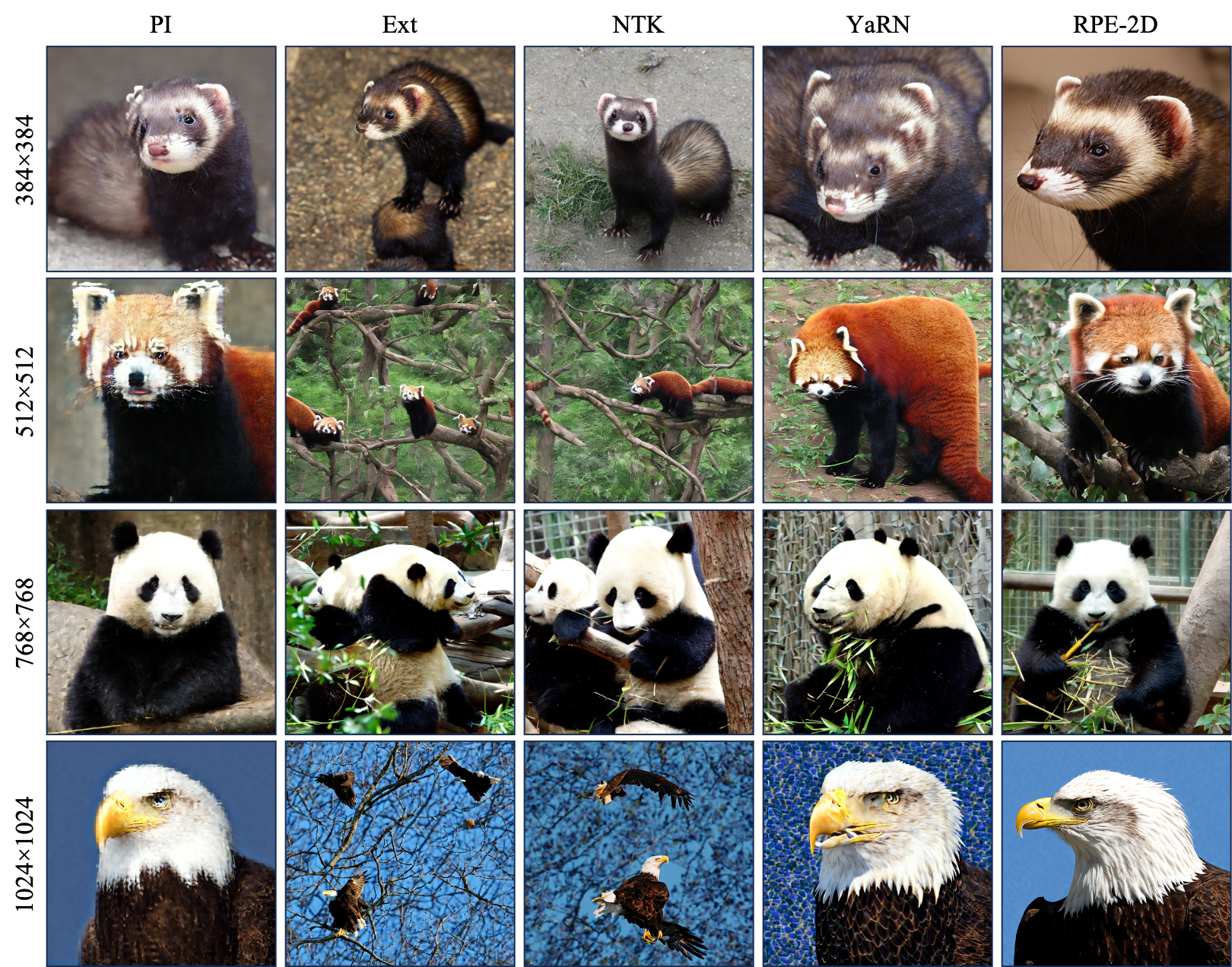}
    \caption{Qualitative results of RPE-2D against different positional encoding extrapolation methods at different resolutions.}
    \label{fig_compare}
\end{figure*}

\subsection{Experimental Setup}
\paragraph{Training Settings} We follow DiT~\cite{peebles2023scalable} using ImageNet-$256^2$ and ImageNet-$512^2$ as training datasets, employing the DiT-XL/2 network architecture while keeping the other training hyper-parameters unchanged\footnote{\url{https://github.com/facebookresearch/DiT}}. On the ImageNet-$256^2$, we trained the model from scratch using the proposed random position encoding for 400k iterations, and compared it with the baseline. Subsequently, we apply the weights obtained from training ImageNet$256^2$ for 400k iterations to ImageNet$512^2$ for an additional 800k iterations, and compared the resolution extrapolation results with the baseline method.

\paragraph{Evaluation Metrics} Following DiT, we use FID~\cite{heusel2017gans}, sFID~\cite{nash2021generating}, IS~\cite{salimans2016improved}, and Precision/Recall~\cite{kynkaanniemi2019improved} as the quantitative 
evaluation metrics in the experiments.
We also follow the default value of 4.0 for ``cfg-scale'' in the \texttt{sample.py} file in the DiT official code.

\begin{table*}[ht]
    \centering
    \begin{tabular}{lcccccccccc}
    \toprule
    \multirow{2}{*}{\centering Method} & \multicolumn{5}{c}{ImageNet $256 \times 256$} & \multicolumn{5}{c}{ImageNet $512 \times 512$} \\
    \cmidrule(lr){2-6} \cmidrule(lr){7-11}
     & FID$\downarrow$ & sFID$\downarrow$ & IS$\uparrow$ & Precision$\uparrow$ & Recall$\uparrow$ & FID$\downarrow$ & sFID$\downarrow$ & IS$\uparrow$ & Precision$\uparrow$ & Recall$\uparrow$ \\
    \midrule
    SinPE & 17.82 & 11.33 & 359.32 & 0.927 & 0.149 & 14.75 & 10.02 & 277.92 & 0.8341 & 0.143 \\
    RoPE & 17.30 & 11.07 & \textbf{366.19} & 0.956 & \textbf{0.176} & 14.17 & 7.16 & 313.73 & \textbf{0.8677} & 0.176 \\
    SinPE w/ RPE-2D & 17.33 & 11.21 & 358.43 & 0.931 & 0.155 & 14.53 & 8.72 & 302.55 & 0.8324 & 0.145 \\
    RoPE w/ RPE-2D & \textbf{16.92} & \textbf{11.02} & 362.14 & \textbf{0.959} & 0.170 & \textbf{14.09} & \textbf{6.61} & \textbf{398.35} & 0.8399 & \textbf{0.225} \\
    \bottomrule
    \end{tabular}
    \caption{Comparison of RPE-2D applied on absolute position encoding (SinPE) and relative position encoding (RoPE).}
    \label{table_comparsion_different_PEs}
\end{table*}

\begin{table*}[ht]
    \centering
    \begin{tabular}{lccccc|ccccc}
    \toprule
    \multirow{3}{*}{\centering Method} & \multicolumn{10}{c}{ImageNet $256 \times 256$} \\
    \cmidrule(lr){2-11}
    & \multicolumn{5}{c|}{ $256 \times 256$ } & \multicolumn{5}{c}{ $512 \times 512$}\\
    & FID$\downarrow$ & sFID$\downarrow$ & IS$\uparrow$ & Precision$\uparrow$ & Recall$\uparrow$ & FID$\downarrow$ & sFID$\downarrow$ & IS$\uparrow$ & Precision$\uparrow$ & Recall$\uparrow$\\
    \midrule
    RPE-2D & 17.17 & 11.19 & 358.17 & 0.9579 & 0.1683 & 20.78 & 27.29 & 293.96 & 0.8155 & 0.155 \\
    + Cond-Aug & 17.09 & 11.13 & 359.55 & 0.9587 & 0.1687 & 19.12 & 23.91 & 325.76 & 0.8399 & 0.163 \\
    + Attention Scale & 17.02 & 11.05 & 361.73 & 0.9578 & 0.1693 & 18.33 & 19.17 & 347.82 & 0.8469 & 0.177 \\
    + Timestep Shift & \textbf{16.92} & \textbf{11.02} & \textbf{362.14} & \textbf{0.9591} & \textbf{0.1703} & \textbf{17.95} & \textbf{18.23} & \textbf{348.99} & \textbf{0.8493} & \textbf{0.181} \\
    \bottomrule
    \end{tabular}
    \caption{Ablation study on components of RPE-2D. Ablation components are progressively integrated in sequence.}
    \label{table_Ablation}
\end{table*}

\begin{figure}
\centering
\includegraphics[width=\linewidth]{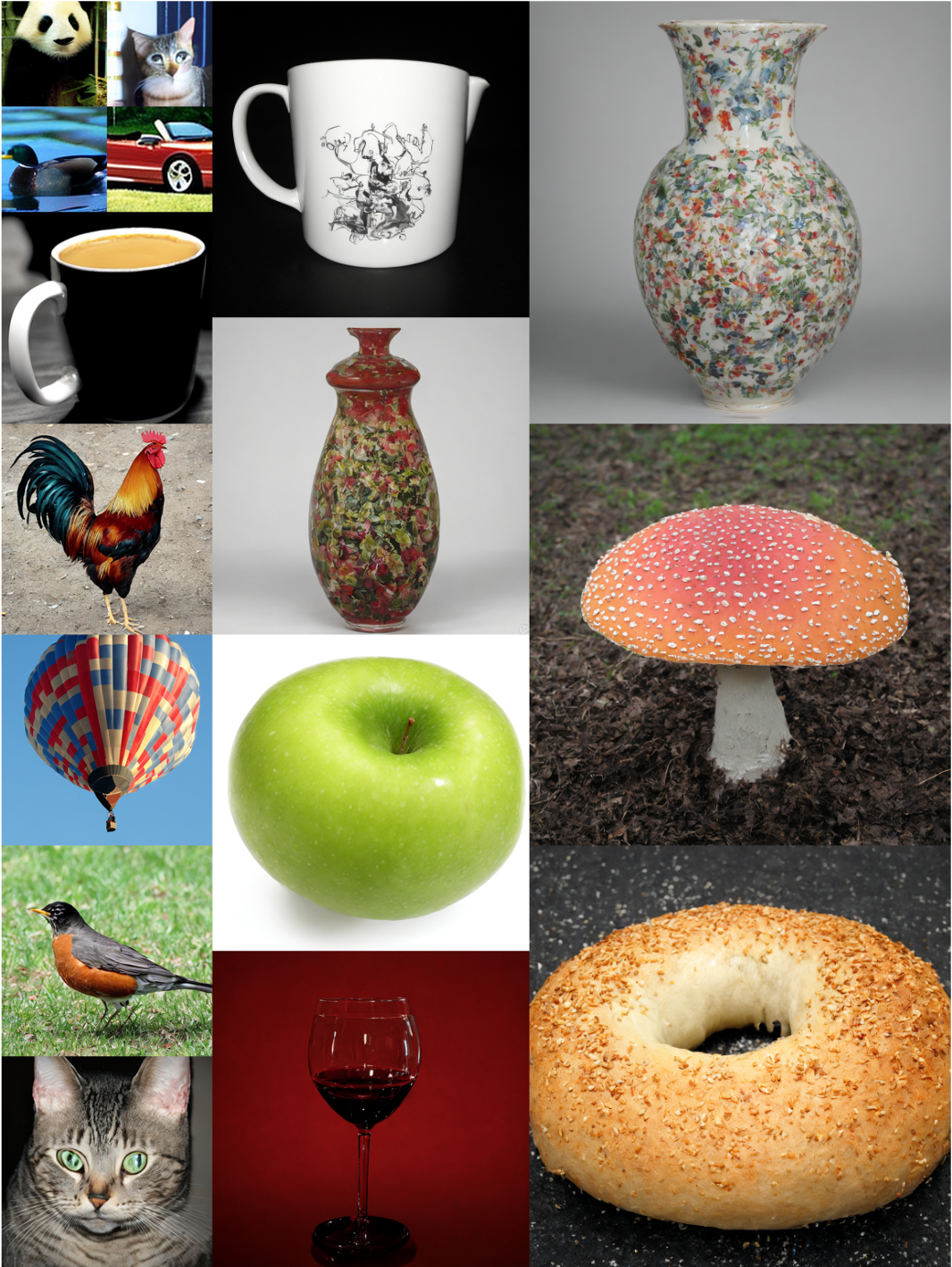}
\caption{Generated images at different resolutions, including $128\times 128$, $256\times 256$, $512\times 512$, $768\times 768$, and $1024\times 1024$, where the model is trained only at resolutions of $256\times 256$ and $512\times 512$.}
\label{fig:resolution}
\end{figure}

\subsection{Comparisons}
RPE-2D is a training approach for positional encodings rather than a specific encoding form, making it theoretically compatible with any type of positional encoding. We apply RPE-2D to both SinPE and RoPE, and the results in~\cref{table_comparsion_different_PEs} show that combining either PE with RPE-2D consistently improves performance.

We then compare RPE-2D with PI~\cite{chen2023extending}, extrapolation (Ext), NTK~\cite{NTK-aware}, and YaRN~\cite{peng2023yarn} for resolution extrapolation, all built on top of RoPE~\cite{su2024roformer}. It is worth noting that all competitors are implemented with two-dimensional positional encodings: in particular, NTK and YaRN are extended to their 2D RoPE-style versions by applying~\cref{eq:RoPE2D} following FiT~\cite{lu2024fit}. Starting from weights trained on ImageNet at $256^2$ for 400k iterations, we extrapolate to $384^2$ and $512^2$. As shown in~\cref{tab:Extrapolation_384_512}, RPE-2D achieves state-of-the-art metrics at both $384^2$ and $512^2$, reducing the previous best sFID of \textbf{34.31} obtained by YaRN to \textbf{18.23}, and thus substantially improving resolution extrapolation. The qualitative results in~\cref{fig_compare} further indicate that RPE-2D maintains superior visual quality at $512^2$, suggesting that our method effectively pushes the practical extrapolation range beyond that of YaRN and NTK. 

We further fine-tune the models on ImageNet at $512^2$ for an additional 800k iterations and extrapolate to $768^2$ and $1024^2$. As reported in~\cref{tab:Extrapolation_768_1024}, RPE-2D continues to obtain the best overall performance at higher resolutions, while PI achieves competitive results on the precision metric. \cref{fig_compare} presents qualitative comparisons between our method and baseline approaches: only RPE-2D consistently preserves both global structure and fine details across resolutions. In contrast, methods such as NTK and YaRN, which combine interpolation and extrapolation, tend to exhibit structural artifacts, whereas PI, as a purely interpolation-based method, often suffers from noticeable detail loss.

\subsection{Ablation Studies} 

We conduct ablation studies on the main components of RPE-2D: (i) the random resize-and-crop augmentation with micro-conditioning (Cond-Aug), (ii) attention scaling, and (iii) timestep shifting.

Our Cond-Aug treats the collected low-resolution training images as resized or cropped views of larger images, where different sampling intervals and starting points correspond to resizing scale factors and cropping coordinates. As shown in~\cref{table_Ablation}, Cond-Aug reduces the FID from \textbf{20.78} to \textbf{19.12} and improves the IS from \textbf{293.96} to \textbf{325.76}, yielding substantial gains at extrapolated resolutions and indicating that it strengthens the modeling of positional order.

In addition, the carefully designed attention scaling and timestep shifting further improve IS while significantly reducing sFID, as reported in~\cref{table_Ablation}, demonstrating their effectiveness when combined with RPE-2D for high-resolution sampling.

\subsection{Applications}

\begin{figure}
\centering
\includegraphics[width=1.0\linewidth]{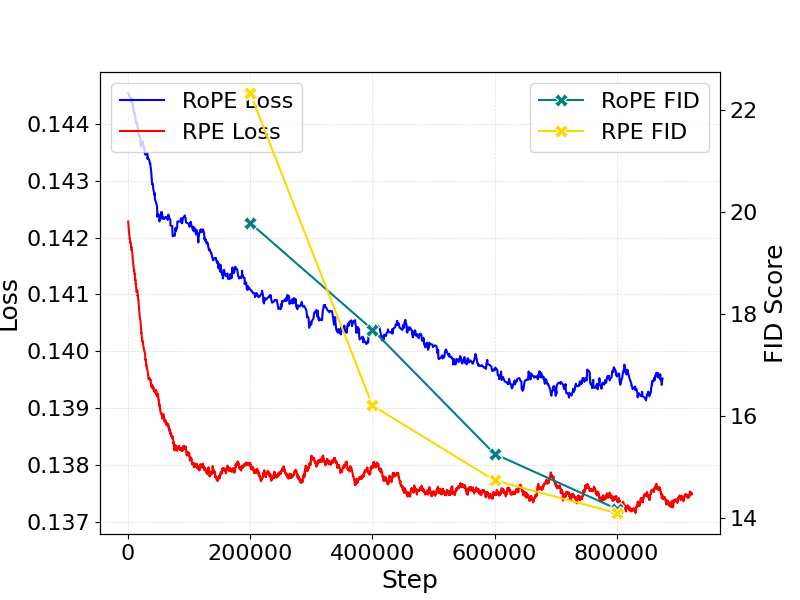}
\caption{Training loss and FID curves of RoPE and RPE-2D.}
\label{fig:loss}
\end{figure}

\subsubsection{Low-Resolution Image Generation}

As shown in~\cref{fig:resolution}, RPE-2D can not only generate images with higher resolutions than those used for training, but also synthesize images at lower resolutions, e.g., generating $128^2$ images when the training resolution is $256^2$. This demonstrates that RPE-2D exhibits resolution generalization in both upward and downward directions.

\subsubsection{Multi-Stage Training Acceleration}
Because RPE-2D enables high-quality high-resolution generation from models trained at lower resolutions, a natural application is to facilitate multi-stage, multi-resolution training. We take a model pre-trained on ImageNet at $256^2$, fine-tune it at a resolution of $512^2$, and compare the loss convergence of standard RoPE against RoPE equipped with our randomized positional encoding scheme. As illustrated in~\cref{fig:loss}, the model with RPE-2D starts from a lower loss and converges more rapidly, which is beneficial for staged training of large-scale models.

%% file: sections/6_conclusion.tex
\section{Conclusion}
This work investigates the resolution generalization problem in diffusion transformers from the perspective of positional encodings (PEs). Previous approaches have not fully addressed the inconsistency of PEs between training and testing. We propose RPE-2D, ensuring that the PEs during testing are all trained. By modeling the position orders among image patches rather than their absolute distances, our method bridge the gap between training and testing. Additionally, we propose random data augmentation further enhance the model's ordering modeling while reducing its dependency on the exact number of tokens. To address the potential issue of image incompleteness caused by random data augmentation, we also introduce micro-conditioning, enabling the model to perceive the specific augmentation methods applied. During high-resolution inference, we also employ attention scaling and timestep shifting to address issues related to attention entropy increase and signal-to-noise ratio mismatch. Experimental results on ImageNet-256/512 demonstrate that our proposed method significantly outperforms existing competing approaches in the resolution generalization problem.